\documentclass[twoside,11pt]{article}
\usepackage{jair, theapa, rawfonts}

\usepackage{amsmath}
\usepackage{multirow}
\usepackage{bbm,url}
\usepackage{graphicx}
\usepackage{amssymb}
\usepackage{booktabs}
\usepackage{adjustbox}
\usepackage{xspace}
\usepackage{algpseudocode}
\usepackage{longtable}
\usepackage{makecell}
\usepackage{pgfplots}

\usepackage{algorithm}
\usepackage{color, colortbl}
\usepackage{float}
\usepackage[caption = false]{subfig}
\usepackage{sectsty}

\usepackage{color}
\usepackage{ulem}
\usepackage{cprotect}
\usepackage{tabularx}
\definecolor{myGreen}{rgb}{0.76,0.93,0.63}
\definecolor{myRed}{rgb}{0.93,0.5,0.5}
\definecolor{LimeGreen}{rgb}{0.2,0.8,0.2}

\jairheading{1}{2022}{1-21}{11/22}{-/-}
\ShortHeadings{ClozE: A Novel Framework for Evaluating the Factual Consistency}
{Li, Li, Hu, Li, Litvak, Natalia \& Zhou}
\firstpageno{1}

\begin{document}

\title{Just ClozE! A Novel Framework for Evaluating the Factual Consistency Faster in Abstractive Summarization}

\author{\name Yiyang Li \email kenlee@bupt.edu.cn \\
        {\it (Co-first author)} \\
       \name Lei Li \email leili@bupt.edu.cn \\
       {\it (Co-first \& Corresponding author)} \\
       \name Dingxin Hu \email hudingxin@bupt.edu.cn \\
       \name Yuze Li \email lyzbupt@bupt.edu.cn\\
       \name Yanquan Zhou \email zhouyanquan@bupt.edu.cn\\
       \it (Corresponding author) \\
       \addr School of Artificial Intelligence\\
       Beijing University of Posts and Telecommunications\\
       Beijing, 100876, P. R. China
       \AND
       \name Marina Litvak \email marinal@ac.sce.ac.il \\
       \name Natalia Vanetik \email natalyav@sce.ac.il \\
       \addr Department of Software Engineering\\ 
       Shamoon College of Engineering \\
       Beer Sheva, 84100, Israel}


\maketitle

\begin{abstract}
The issue of factual consistency in abstractive summarization has received extensive attention in recent years, and the evaluation of factual consistency between summary and document has become an important and urgent task. Most of the current evaluation metrics are adopted from the question answering (QA) or natural language inference (NLI) task. However, the application of QA-based metrics is extremely time-consuming in practice while NLI-based metrics are lack of interpretability. In this paper, we propose a cloze-based evaluation framework called ClozE and show the great potential of the cloze-based metric. It inherits strong interpretability from QA, while maintaining the speed of NLI-level reasoning. We demonstrate that ClozE can reduce the evaluation time by nearly 96$\%$ relative to QA-based metrics while retaining their interpretability and performance through experiments on six human-annotated datasets and a meta-evaluation benchmark GO FIGURE \shortcite{gabriel2021go}. Finally, we discuss three important facets of ClozE in practice, which further shows better overall performance of ClozE compared to other metrics. The codes and models are released at https://github.com/Mr-KenLee/ClozE .
\end{abstract}

\section{Introduction}
\label{sec:Introduction}
The factual consistency problem in abstractive summarization is a major challenge that limits the application of the summaries generated by state-of-the-art models in practice. A number of works \shortcite{cao2018faithful,maynez2020faithfulness,deutsch2021understanding} have shown that even the best abstractive summarization models still suffer from many factual inconsistencies. Thus, determining the factual consistency between a summary and its document is critical to addressing this challenge.

\begin{table*}[t]\small
\centering{
\begin{tabular}{c|c|c}
\hline
\multicolumn{3}{c}{\textbf{Document $\&$ Summary}} \\
\hline
\multicolumn{3}{l}{\textbf{D:} England coach peter moores talks to \textbf{\textcolor{LimeGreen}{the news media}} during a press conference at the ade-}\\
\multicolumn{3}{l}{laide oval on sunday.} \\
\multicolumn{3}{l}{\textbf{S:} Peter moores talks to \textbf{\textcolor{red}{the adelaide oval}} on sunday.} \\
\hline
\hline
\textbf{NLI-based} & \textbf{QA-based} & \textbf{Cloze-based}\\
\hline
\textbf{Entailment Matrix} & \textbf{Extract Factual Factor} & \textbf{Extract Factual Factor}\\
\multirow{5}{*}{
$   \begin{pmatrix}
		\text{Entail.} \\
		\text{Neutral} \\
		\text{Contra.}
	\end{pmatrix}
	=
	\begin{pmatrix}
		0.2 \\
		0.1 \\
		0.7
	\end{pmatrix}
	$
} 
& the adelaide oval & the adelaide oval\\
                            \cline{2-2}\cline{3-2}
                            & \textbf{Question Generation (QG)} & \textbf{Masking Summary}\\
                            & Who does Peter talk to? & Peter moores talks to \verb|[M]| on sunday.\\
                            \cline{2-2}\cline{3-2}
                            & \textbf{Question Answering (QA)} & \textbf{Filling in the blank} \\
                            & the news media & the news media \\
\hline
\textbf{Entailment Score} & \textbf{Similarity Score} & \textbf{Similarity Score} \\
0.2                  & 0.4                            & 0.4 \\

\hline

\end{tabular}}
\caption{A simple example of three categories of metrics. NLI-based metrics: The factual consistency score is computed with the probability of the summary being entailed by the document. QA-based metrics: (1) Extracting factual factors from the summary as answers. (2) Generating associated questions through the QG module. (3) Answering the questions based on the document through the QA module. (4) Computing the score based on the similarity between the extracted factual factors and generated answers.  Cloze-based metrics (ClozE): The overall process are similar to QA-based metrics, with the difference that the QG module is replaced with a model-free strategy. The factual factors in the summary are masked at first and then filled by a cloze model.}
\label{tab:1}
\end{table*}

In recent years, researchers have proposed some abstractive summarization evaluation metrics for factual consistency, which can be classified mainly into two categories: NLI-based metrics \shortcite{falke2019ranking,kryscinski2020evaluating,laban2022summac} and QA-based metrics \shortcite{wang2020asking,durmus2020feqa,scialom2021questeval}. We provide a simple comparison and description for them in Table \ref{tab:1}.

As shown in Table \ref{tab:1}, due to the one-stage computation with NLI modeling, NLI-based metrics have faster evaluation speed but poorer interpretability. They cannot locate the factual inconsistency, which is also an essential task in analyzing factual consistency. In contrast, QA-based metrics have strong interpretability based on the explicit definition of facts. However, their multi-stage computational process results in large computational overheads \shortcite{koto2022ffci}. Furthermore, QA-based metrics are calculated in several independent modules, which results in accumulated errors. For example, the QA module fails to generate answers to roughly 14$\%$ of the questions that can be answered \shortcite{durmus2020feqa}, and questions generated by the QG module need to be carefully filtered. 

Faced with the pros and cons of the two categories of metrics, we focus on the cloze task \shortcite{taylor1953cloze} and propose a cloze-based metric ClozE (\textbf{Cloz}e-based Factual Consistency \textbf{E}valuation), which is a fast and lightweight factual consistency evaluation framework that leverages the best of both NLI-based and QA-based metrics while avoiding their limitations.  To the best of our knowledge, ClozE is the first cloze-based metric for evaluating factual consistency. Specifically, our contributions are the following:

(1) In order to maintain the interpretability, we leverage the extracting-comparing-facts framework in QA-based metrics. But instead of generating questions and answers with QG and QA modules, we simply use a cloze module, where the questions are defined as text spans whose factual factors are replaced with \verb|[MASK]| and the cloze results are answers. This one-stage modeling can compute faster than QA-based metrics while being more interpretable than NLI-based ones. Meanwhile, it can maximize the use of knowledge from masked language modeling (MLM) based pre-trained language models (PLMs) due to the similarity between the upstream and downstream tasks.

(2) Experiments on two benchmarks show that ClozE can achieve a comparable correlation with human judgments as QA-based and NLI-based metrics.  Compared to QA-based metrics, ClozE reduces the time consumption by 96$\%$ on average and obtains higher accuracy in generating answers, 30$\%$ above QA-based metrics. Meanwhile, ClozE is not only more interpretable but also more accurate for more abstract summaries than NLI-based metrics.

(3) We further analyze three important facets of ClozE in practice, including the balance of speed and performance,  optimization of metric and location of factual errors. We demonstrate that ClozE performs better than QA-based and NLI-based metrics on these three aspects combined, fully showing the usefulness and great potential of our proposed framework.

\section{Related Work}
\label{sec:Related Work}

\subsection{Non-factual metrics}
N-gram overlap-based metrics were generally used for summarization before the concept of factual consistency was proposed, such as ROUGE \shortcite{lin2004rouge}, BLEU \shortcite{papineni2002bleu} and METEOR \shortcite{banerjee2005meteor}. Recent work has applied pre-trained models to evaluate summaries. BERTScore \shortcite{zhang2019bertscore}, BLANC \shortcite{vasilyev2020fill} and COMET \shortcite{rei2020comet} give scores by calculating the similarity between a candidate summary and reference summaries with Bert \shortcite{kenton2019bert}. \shortciteA{sellam2020bleurt} proposes BLEURT, which pre-trains a fully learned metric on large amounts of synthetic data and fine-tunes it on human ratings.

\subsection{NLI-based metrics}
\shortciteA{falke2019ranking},\shortciteA{barrantes2020adversarial} and \shortciteA{kryscinski2020evaluating} trained an NLI model to score candidate summaries for factual consistency and obtained a high correlation with human judgments. \shortciteA{laban2022summac} divided a document and a summary into multiple blocks and applied the NLI model to calculate the blocks for an entailment matrix to resolve the problem of the granularity mismatch. 

\subsection{QA-based metrics}
SummaQA \shortcite{scialom2019answers}, QAGS \shortcite{wang2020asking}, FEQA \shortcite{durmus2020feqa} and QuestEval \shortcite{scialom2021questeval} are all QA-based metrics which used a similar process as Table \ref{tab:1} to obtain the final score by comparing the extracted summary with the factual factors of the document. \shortciteA{nan2021improving} proposed QUALS, which greatly improves the speed of Q\&A by allowing the model to generate questions and answers simultaneously. \shortciteA{fabbri2021qafacteval} have experimented and optimized each module of the QA-based methods to achieve better performance than other QA-based metrics.

\subsection{Factual Consistency Benchmark}
Assessing the validity of factual consistency metrics is also a challenging task. In recent years, more work has given relevant fact consistency annotated datasets as well as meta-evaluation methods. \shortciteA{wang2020asking,fabbri2021summeval} separately proposed a crowdsourced annotated dataset on the factual consistency, as a way to evaluate the correlation of factual consistency metrics to human judgments. \shortciteA{laban2022summac} fused previous works and standardize their datasets. Meanwhile, the Pearson correlation coefficient is commonly used to evaluate the correlation between the above metrics and human judgments. 
In contrast, \shortciteA{gabriel2021go,pagnoni2021understanding} proposed the meta-evaluation of factual consistency with more refined criteria for the performance evaluation of factual consistency metrics. 

\section{Methods}
\label{sec:Methods}
\begin{figure*}
    \includegraphics[width=\textwidth]{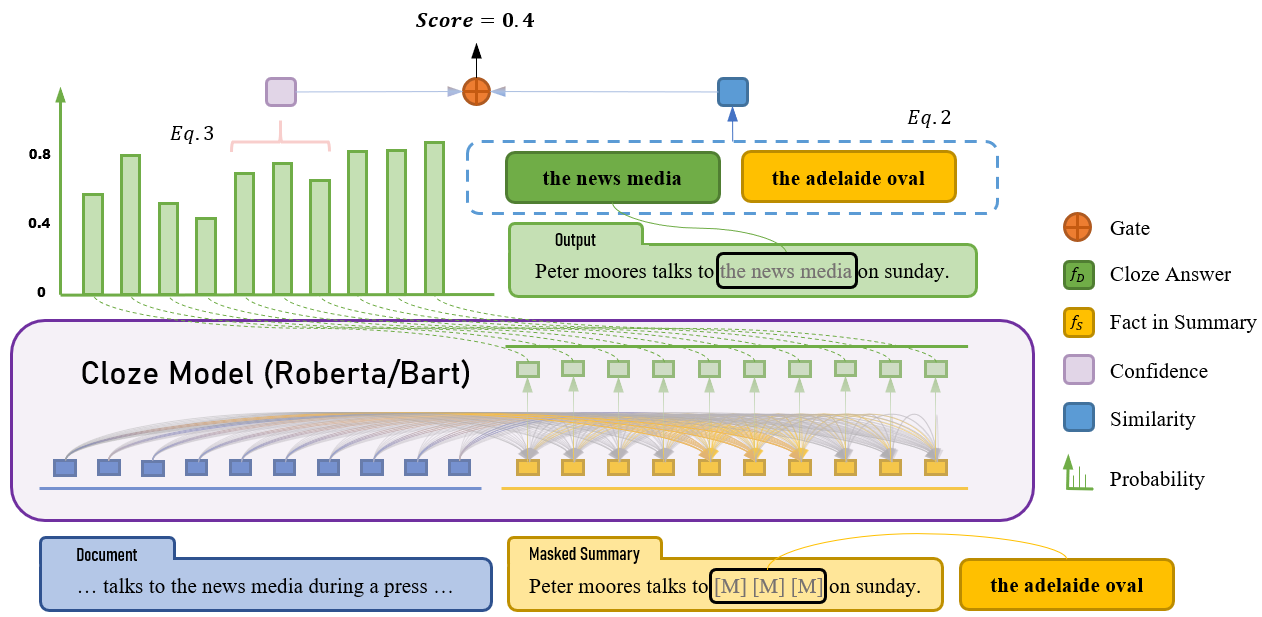}
\cprotect\caption{We show the overall framework of ClozE. We first extract all factual factors $f_s$ according to the summary and replace $f_s$ with \verb|[MASK]| identifiers at the token-level in the summary (\verb|[M]| part of the Summary in the figure). By inputting the masked summary into the cloze model together with the document, we can later obtain the cloze answers $f_D$ corresponding to the document at the position previously replaced with \verb|[MASK]|. Finally, we use the equations given in Eq.\ref{equation:2} and Eq.\ref{equation:3} to calculate the final ClozE score.}
\label{figure:1}
\end{figure*}

We now introduce our ClozE for factual consistency evaluation. 
Firstly, the overall framework of ClozE is shown in Table \ref{tab:1} and Figure \ref{figure:1}. 
In the next subsections, we introduce the modules in our ClozE framework according to a practical evaluation process. In order to show the great potential and application value of our framework, the modules in our framework are very simple and basic.

\subsection{ClozE}
The modeling approach in QA-based metrics is naturally interpretable so that we adopted its framework\footnote{Extract factual factors and compare them.} . In order to alleviate the problems existing in QA-based metrics which are mainly caused by the multi-module collaboration, we consider merging the QG and QA parts so that only one model will be computed under the whole framework. 

Following the motivation, we introduce cloze task, an elegant but lightweight modeling method. We transform a question into a summary text with $k$ factual factors masked off and the answer is generated by the cloze model. In this case, since questions can be obtained directly, we can omit the QG module and substitute the QA module with a cloze model. We present a formula similar to the one proposed by \shortciteA{wang2020asking}.

Given the document $X$, the summary $Y$ and the masked summary (cloze question) $Y^{'}$, each of which consists of a series of tokens, we can obtain $P(A|X,Y^{'})$ and $P(A|Y)$, which respectively represent the cloze answers generated by the cloze model and the factual factors in the summary $Y$. Therefore, the factual consistency score can be computed by the following formula.
\begin{equation}
{E_{A\sim{}P(A|Y)}[D(P(A|X,Y^{'}),P(A|Y))]} 
\end{equation}
where $D(*,*^{'})$ denotes a function that measures the distribution of both $*$ and $*^{'}$.

Due to the similarity to MLM, the cloze model in ClozE can be trained by prompt learning\shortcite{liu2021pre} and maximize the use of knowledge in PLMs. Meanwhile, it is also a self-supervised task and does not require annotated datasets, which prevents from severe data bias\shortcite{cao2020factual,goyal2021annotating} caused by data augmentation similar to FactCC\shortcite{kryscinski2020evaluating}. 
Besides, it still maintains the same interpretability and is capable of factual error correction in some level.

\subsection{Factual Factors}
In the work of \shortciteA{pagnoni2021understanding}, factual errors shaped as named entities and noun phrases account for the majority of all factual errors, and both fit well in line with the task of cloze in our framework. As a result, we define named entities and noun phrases as factual factors in our framework.

\subsection{Cloze Model}
\label{sec:clozemodel}
\begin{figure}
    \begin{center}
        \includegraphics[width=4in]{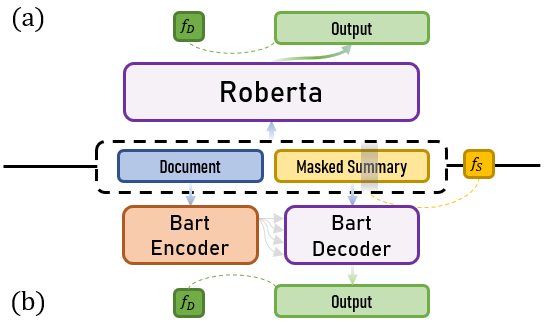}
    \end{center}
    \caption{Two kinds of cloze models are used for the ClozE. The first one is Roberta (a), where we concatenate the document and the masked summary as the input of the model. The second one is Bart (b), where we take the document as the Encoder input and the masked summary as the Decoder input.}
    \label{figure:2}
\end{figure}

Cloze models occupy the key position in our work. For fairness considerations of most previous metrics (both QA-based and NLI-based metrics), we conduct experiments using two types of basic pre-trained models, Roberta \shortcite{liu2019roberta} and Bart \shortcite{lewis2020bart}, as shown in Figure \ref{figure:2}. 

The approach using Roberta is a general form of our framework. The pre-training task of Roberta is a better match to the cloze task, where we concatenate the document and the masked summary together by feeding them into the model, expecting the model output to be a complete text.

We also use the Bart model because of its training and prediction runtimes. In our task, we actually do not require the model to regenerate the part of the document, so this part can be combined with the masked summary by cross-attention mechanism\footnote{It only computes the attention scores between the decoder hidden states and the encoder outputs in the decoding stage.} instead of self-attention mechanism \shortcite{vaswani2017attention}. It is important to state that we leverage MLM task in the decoder instead of the original autoregressive task in Bart. This setting is intended to facilitate comparison with the Roberta-based ClozE model.

\subsection{The ClozE Score}
\label{sec:k}
In line with the approach adopted in the QA-based metrics in recent years, we use token-level F1 as the quality for the comparison of factual factors. Because F1 focuses on whether a token appears in both two sequences, it is not sensitive to errors such as misalignment and missing tokens. This characteristic can help to mitigate the errors caused by the performance of the cloze model to some extent.

For a summary with $N$ factual factors, we extract all of the factual factors and split them into $\lceil N/k \rceil$ groups, where $k$ is the number of factual factors for a single masking. Therefore, a summary will need to be calculated $\lceil N/k \rceil$ times by the cloze model, and the final ClozE score is computed using the following equations:
\begin{equation}
\label{equation:2}
Score=\frac{\sum_{i=1}^{\lceil N/k \rceil} \sum_{j} F1(G_{ij},S_{ij})}{N}
\end{equation}
where $F1(*,*^{'})$ denotes the function which computes the $F1$ score between the two symbols in the tuple. $G_{ij}$ and $S_{ij}$ respectively denote the factual factor in the summary and generated with the cloze model.

\subsection{Granularity}
Most of the factual consistency metrics calculate the factual consistency score with the sentence-level units. Out of preciseness, we take the summary level into account. We concatenate one document with one masked sentence instead of a full masked summary. This means that a masked summary is divided into multiple sentences as the inputs for cloze model.  Afterwards, the multiple cloze answers are collected and the evaluation scores are calculated uniformly using Eq.\ref{equation:2}. We also conduct experiments with both levels. 

\subsection{Confidence}
\label{sec:ab}
We take into account the confidence mentioned in the work of \shortciteA{scialom2019answers} and \shortciteA{qasemi2022evaluating}. We propose a convenient and effective way to calculate the confidence by Eq.\ref{equation:3}, which is the average prediction probability of the tokens contained by the masked factual factors. 

\begin{equation}\label{equation:3}
    Confidence =\frac{1}{|T_{f}|} \sum_{t\in{T_{f}}}p(t)
\end{equation}
where $t$ denotes a token in $T_{f}$ while $T_{f}$ denotes all the tokens contained in factual factor $f$.

In addition, we introduce two thresholds $\alpha$ and $\beta$. When the confidence is less than $\alpha$ and the current ClozE score is less than $\beta$, the ClozE score will be set to 0. This trick is designed to make ClozE more robust by ignoring some of the score increases due to incorrectly filled blanks.

\section{Experiments}
\label{sec:Experiments}

\subsection{Baseline Metrics}
Since there has been a lot of previous work\shortcite{wang2020asking,scialom2021questeval,fabbri2021summeval,laban2022summac} that has demonstrated the inability of model-free metrics to effectively measure factual consistency, we only consider model-based metrics as our baselines.

\vspace*{-0.05in}
\paragraph{Non-Factual Baselines} 
We use BERTScore and BLANC-help to verify the sensitivity of the proposed method for factual consistency. We do not perform speed tests on the above metrics because they do not address the factual consistency issue. 

\vspace*{-0.05in}
\paragraph{Factual Baselines} We use QAGS, FEQA, SummaQA, QUALS, QuestEval and QAFactEval \shortcite{fabbri2021qafacteval} as QA-based baselines and FactCC \shortcite{kryscinski2020evaluating}, ANLI \shortcite{barrantes2020adversarial} and SUMMAC \shortcite{laban2022summac} as NLI-based baselines. We verify that our method performs comparable with these factual consistency metrics while avoiding their limitations through experiments.

\vspace*{0.2in}
We test most of the baselines above with the code and models they provided for fairness. For ANLI, we used the results from \shortciteA{fabbri2021qafacteval} because of the lack of its code and model. Besides, due to some problems with the official QAGS code, we opted for the factsumm\footnote{\url{https://github.com/Huffon/factsumm}} implementation for QAGS. In addition, we do not use any acceleration strategies such as multi-processing and half-precision inference for all baselines. 

\subsection{Factual Evaluation Benchmark}
We use two benchmarks to evaluate the baselines and our framework, including Pearson Benchmark and GO FIGURE Benchmark. We believe that more meta-evaluation will give a more objective evaluation for these metrics. In this section, we briefly describe these two benchmarks while providing a more detailed description in Appendix A.

\subsubsection{Pearson Benchmark} 
Following \shortciteA{fabbri2021qafacteval}, we use six human-annotated datasets to evaluate factual consistency metrics with Pearson correlation coefficient. The summaries used for evaluation are generated by a variety of different models on CNN/DM and XSum \shortcite{narayan2018don}, respectively. These datasets are XSumFaith (XSF)\shortcite{maynez2020faithfulness}, SummEval (SE)\shortcite{fabbri2021summeval}, FRANK\shortcite{pagnoni2021understanding} and QAGS\shortcite{wang2020asking}. We partition FRANK and QAGS into FRANK-CNN/DM (FCD), FRANK-XSum (FXS), QAGS-CNN/DM (QCD) and QAGS-XSum (QXS), resulting in a total of six test datasets.

\subsubsection{GO FIGURE Benchmark} In addition, as suggested in \shortciteA{vasilyev2020fill}, human scores are error-prone and being correlated with human scores should not be considered as a sufficient and unique validation, but rather as an independent confirmation of the reasonableness of the metrics. Therefore, we use GO FIGURE \shortcite{gabriel2021go}, which has developed a systematic criterion for factual consistency metrics, as another benchmark to evaluate baselines and our work from a different aspect. 

\subsection{Experimental Details}
\label{sec:details}

\subsubsection{Extractor} We use SpaCy \footnote{\url{https://spacy.io/api}} to extract named entities and noun phrases. We first used \verb|en_core_web_sm|, the model adopted in previous QA-based metrics work, to extract factual factors. After that, we also adopted the \verb|en_core_web_trf| model with higher accuracy to extract factual factors. \verb|en_core_web_sm| and \verb|en_core_web_trf| are both models used in SpaCy for text analysis, the former is designed for cpu only and the latter is based on Roberta's extraction model.

By using SpaCy, we obtained too many noun phrases, and some of them have overlapping parts with named entities. Therefore, we take named entities as the main part and remove those noun phrases containing overlapping parts, to get a set of factual factors that do not interfere with each other. 

\subsubsection{Cloze Model} We use transformers library \shortcite{wolf2019huggingface} to build and employ  Roberta-base and Bart-base pre-trained models. We denote the ClozE based on Roberta-base model and Bart-base model as ClozE-R and ClozE-B respectively.

\subsubsection{Cloze Task Dataset}
Since the input of our cloze model is a document and its summary, it is a good choice to use the summarization dataset to train the cloze model. And we believe that the golden summaries in the summarization dataset have high factual consistency. Due to this, we adopt the summarization dataset CNN/DM\shortcite{hermann2015teaching} as the dataset for our cloze task.
We use their training set as the training set for our cloze task and their test set to evaluate the performance of our cloze model.

\subsubsection{Training Settings} 
The input lengths of the models are limited to 512. For every training sample, we first select a random sentence in the summary and randomly mask a number of factual factors in that sentence for cloze task. Cross entropy loss with Adam \shortcite{kingma2014adam} optimizer is used, with a batch size of 96, a learning rate of 1e-4 and a warming up step of 8000. We trained and tested our models on two NVIDIA GeForce RTX 3090 GPUs for 2-3 days with the parameters $k = 1$ (\S \ref{sec:k}) and $\alpha=\beta=0.5$ (\S \ref{sec:ab}).

\begin{table*}[t]
    \begin{center}
        \begin{adjustbox}{max width=1\linewidth}
\begin{tabular}{clccccccccc}
\hline
\textbf{Type} & \textbf{Metric} &\textbf{XSF} &\textbf{SE} &\textbf{FCD} &\textbf{FXS} &\textbf{QCD} &\textbf{QXS} & \textbf{Overall} & \textbf{Sec./Sum} & \textbf{Inc.}\\
\hline
\multirow{2}{*}{NF} & BERTScore & 20.00 & 24.19 & 41.69 & 19.26 & 57.19 & 1.83 & 27.36 & \multirow{2}{*}{/} & 1 \\
                    & BLANC-help & 10.94 & 17.08 & 23.68 & 12.22 & 18.50 & 4.01 & 14.41 &  & 1\\
\hline
\multirow{4}{*}{NLI} & FactCC & 4.78 & 42.11 & 51.87 & 8.15 & 57.63 & 34.16 & 33.12 & \textbf{0.06} & 1 \\
                     & ANLI* & 16.00 & 43.00 & 53.00 & 18.00 & 65.00 & \textbf{\uuline{39.00}} & 39.00 & / & 1 \\
                     & SUMMAC & 6.93 & \textbf{\uline{49.67}} & \textbf{\uuline{56.19}} & \textbf{\uline{32.24}} & 57.30 & \textbf{\uline{44.01}} & 41.06 & 0.19 & 2 \\
\hline
\multirow{6}{*}{QA} & SummaQA & 14.82 & 14.69 & 16.14 & 6.19 & 28.28 & 9.90 & 15.00 & 3.39 & 1\\
                    & FEQA & 29.44 & 43.71 & 52.50 & 14.96 & 55.81 & 31.72 & 39.28 & 3.00 & 2\\
                    & QAGS(factsumm impl.) & 11.90 & 20.32 & 26.06 & 10.64 & 22.65 & 22.22 & 18.31 & 2.20 & 2 \\
                    & QUALS & \uline{36.81} & 27.32 & 49.21 & 20.32 & 56.97 & 19.47 & 35.01 & 1.01 & 1 \\
                    & QuestEval & \textbf{46.38} & 37.75 & 52.66 & \textbf{\uuline{30.05}} & 53.39 & 27.81 & \textbf{\uline{44.05}} & 4.37 & 2 \\
                    & QAFactEval(lerc) & \textbf{\uuline{33.93}} & \textbf{61.06} & \textbf{67.60} & \textbf{38.42} & \textbf{71.66} & \textbf{49.03} & \textbf{54.53} & 1.96 & 3\\
                    \hline
\multirow{4}{*}{ClozE-R} & \textbf{Base} & 16.45 & 44.29 & 49.70 & 21.33 & \textbf{\uuline{65.76}} & 27.79 & 39.50 & 0.13 & 1\\
                        & \verb| +sentence-level| & 16.45 & 47.17 & \textbf{\uline{57.27}} & 21.27 & 65.20 & 27.79 & 41.47 & 0.11 & 1\\
                        & \verb| +en_core_web_trf| & 18.66 & 47.09 & 55.38 & 23.10 & \textbf{\uline{66.12}} & 32.36 & 42.07 & 0.23 & 1\\
                        & \verb| +confidence| & 19.22 & \textbf{\uuline{48.01}} & 55.12 & 23.54 & 65.59 & 29.37 & \textbf{\uuline{42.30}} & 0.23 & 1\\\hline
                    
\multirow{4}{*}{ClozE-B} & \textbf{Base} & 20.05 & 38.17 & 45.21 & 19.31 & 63.39 & 20.56 & 37.23 & \textbf{\uuline{0.09}} & 1\\
                        & \verb| +sentence-level| & 20.08 & 35.33 & 42.12 & 19.27 & 61.07 & 20.56 & 25.57 & \textbf{\uline{0.08}} & 1\\
                        & \verb| +en_core_web_trf| & 20.44 & 36.82 & 43.26 & 20.40 & 61.68 & 19.42 & 36.52 & 0.19 & 1\\
                        & \verb| +confidence| & 20.44 & 36.47 & 42.12 & 20.89 & 61.01 & 21.99 & 36.18 & 0.19 & 1\\ \hline
\end{tabular}
\end{adjustbox}
\end{center}
\cprotect\caption{Pearson correlation coefficients of each metric relative to human judgments, the time spent on the evaluation (\textbf{Sec./Sum}) and the inconvenience of metrics (\textbf{Inc.}). The coefficients under different signatures separately denote the \textbf{best}, \textbf{\uline{second}} and \textbf{\uuline{third}} ones. \textbf{Sec./Sum} means the number of seconds it takes to evaluate a summary. The table also shows the results of the ablation study in the Section ClozE-R and ClozE-B, where \textbf{Base} denotes \verb|en_core_web_sm+summary-level| setting.} 
\label{tab:2}
\end{table*}

\subsection{Results and Analysis}
\label{sec:results}

For objectivity and comprehensiveness, we evaluate ClozE and baselines in multiple aspects of performance, time consumption, interpretability and convenience. We emphasize that all these aspects are equally important and our main purpose is to propose a new factual consistency evaluation framework with a strong potential rather than pursuing all aspects of state of the arts (SOTA).

\subsubsection{Pearson Benchmark}
We show the Pearson correlation coefficients between human-annotated judgements and various metrics in Table \ref{tab:2}. To facilitate the overall analysis, we follow \shortciteA{laban2022summac} and compute the average as the overall performance on the benchmark. For all the datasets, the correlation scores are sufficiently plausible with $p<0.01$.

Our framework ranks 3rd out of 11 baselines by overall scores on six datasets. And for multiple benchmarks, our framework has a high ranking compared to the previous methods. Though we do not obtain the best performance, the results also show that our framework has 
a strong potential and a comparable performance with other framework metrics.

We also analyze the gap between our framework and the better ones. 
Compared with QuestEval and QUALS, our framework performs worse on XSF while better on the other datasets. It indicates that we are required to pay more attention to the special data distribution on XSF.  
And compared with QAFactEval, the scores of our framework are lower than theirs. We believe that the better performances of QAFactEval are due to their optimization for each module in QA-based metric, while our ClozE only uses the most basic PLMs. We will carry out further optimization work afterward.  

And under different cloze models, ClozE-B does not perform as well as ClozE-R. The reason is that the Bart decoder uses a causal mask mechanism, which makes the model unable to perceive the following information of the summary text and leads to the inability to fill in some blanks correctly. 

\subsubsection{GO FIGURE Benchmark}
We present the results for each baseline under GO FIGURE benchmark in Table \ref{tab:3} and Table \ref{tab:4}. We analyze the GO FIGURE results in terms of three aspects, including factual sensitivity, lower bound and upper bound.

\paragraph{Factual Sensitivity} Factual sensitivity is mainly evaluated by Correlation and p-value. The results show that all baselines and ClozE are negatively correlated with the factual error level and the p-values are small enough to be credible on both CNN/DM and XSum datasets. It indicates that ClozE is sensitive to both entity and non-entity factual errors, which is consistent with the existing factual consistency evaluation metrics.

\paragraph{Lower Bound} The "Lower Bound" in GO FIGURE represents the random text, which is not relevant to the document. We find only FactCC gains a higher score in "Lower Bound" than that in "Upper Bound" on XSum, which is a bad case for factual consistency metrics. It shows that FactCC is not robust on more abstractive text.

\paragraph{Upper Bound} "Upper Bound" in GO FIGURE represents the summary text without factual errors, which is the golden summary actually. This means that for extracting-comparing-facts metrics like QA-based metrics and ClozE, the scores they give to "Upper Bound" is the accuracy of the factual factors they generate. The results show that most of QA-based metrics generate low-accurate answers while ClozE-R can generate more accurate ones (\textbf{+30$\%$} on average). In addition, the scores given by baselines on XSum are much lower than those on CNN/DM while ClozE maintains a better performance, which shows that our framework is better and more robust while evaluating abstract texts.

\subsubsection{Time Consumption}
We use QAGS-CNN/DM to measure the time consumption on the same device. As shown in Table 2, our method has an extremely significant improvement over the QA-based metrics and comparable with NLI-based metrics, which reduces the time consumption by nearly 96$\%$ on average compared to QA-based metrics. This great improvement is mainly due to : (1) we use a smaller model compared to the models in most of QA-based metrics, and (2) most importantly, we discard the most time-consuming QG module, which greatly reduces the runtimes for evaluation. 

We also observe that ClozE-R takes nearly twice the evaluation time of ClozE-B, whose reason is mentioned in \S \ref{sec:clozemodel}.

\begin{table*}[t]
\begin{center}
\begin{adjustbox}{max width=1\linewidth}
\begin{tabular}{l|cc|ccccc|cc}
\hline
 & \textbf{FactCC} &\textbf{SUMMAC} &\textbf{SummaQA} &\textbf{FEQA} &\textbf{QUALS} &\textbf{QuestEval} &\textbf{QAFactEval(f1)} &\textbf{ClozE-R} &\textbf{ClozE-B}\\
\hline
Upper Bound ($\%$) & 44.04 & 27.14 & \colorbox{myRed}{10.87} & \colorbox{myRed}{37.56} & \colorbox{myRed}{-75.55} & \colorbox{myRed}{34.63} & \colorbox{myRed}{46.63} & \colorbox{myGreen}{70.10} & \colorbox{myRed}{48.90}\\
Level 1 ($\%$)& 36.44/38.71 & 16.07/11.28 & 9.58/10.63 & 33.35/36.64 & -86.26/-82.43 & 31.92/32.31 & 37.27/37.70 & 64.80/69.28 & 45.58/48.44\\
Level 2 ($\%$)& 32.30/33.80 & 10.70/0.30 & 8.53/10.38 & 28.46/36.13 & -99.83/-87.86  & 30.41/30.67 & 31.66/32.07 & 61.00/69.05 & 43.14/47.97\\
Level 3 ($\%$)& 29.58/28.89 & 2.99/-9.99 & 7.54/10.26 & 25.12/35.63 & -107.22/-92.61 & 28.45/29.21 & 27.34/25.79 & 57.16/68.40 & 40.38/47.60\\
Lower Bound ($\%$)& 22.29 & -18.87 & 0.12 & 0.58 & -373.01 & 0.68 & 0.01 & 18.77 & 7.38\\
\hline
\colorbox{myGreen}{Correlation} & -0.99/-1.00 & -0.99/-1.00 & -1.00/-0.98 & -0.99/-1.00 & -0.98/-1.00 & -1.00/-1.00 & -0.99/-0.99 & -1.00/-0.96 & -1.00/-1.00\\
\colorbox{myGreen}{p-value} & 0.08/$<$0.01** & 0.07/0.01* & 0.01*/0.13 & 0.07/$<$0.01** & 0.11/0.02* & 0.04*/0.02* & 0.05*/0.02* & $<$0.01**/0.17 & 0.02*/0.04*\\
\hline
\end{tabular}
\end{adjustbox}
\end{center}
\caption{Results of various metrics under the GO FIGURE benchmark on CNN/DM. Upper and Lower Bound denote the scores given by the metrics under golden summaries and random texts respectively. Level * denotes the scores given by the metrics under error level *. Correlation denotes the correlation between the metrics and error level. In each cell (*/*), entity factual errors are on the left and non-entity factual errors are on the right. Results marked \colorbox{myRed}{red} and \colorbox{myGreen}{green} respectively denote bad and good cases that deserve attention.}
\label{tab:3}
\end{table*}

\begin{table*}[t]
\begin{center}
\begin{adjustbox}{max width=1\linewidth}
\begin{tabular}{l|cc|ccccc|cc}
\hline
 & \textbf{FactCC} &\textbf{SUMMAC} &\textbf{SummaQA} &\textbf{FEQA} &\textbf{QUALS} &\textbf{QuestEval} &\textbf{QAFactEval(f1)} &\textbf{ClozE-R} &\textbf{ClozE-B}\\
\hline
Upper Bound ($\%$) & \colorbox{myRed}{22.40} & \colorbox{myRed}{-33.35} & \colorbox{myRed}{4.48} & \colorbox{myRed}{27.87} & \colorbox{myRed}{-141.69} & \colorbox{myRed}{19.95} & \colorbox{myRed}{14.70} & \colorbox{myGreen}{65.00} & \colorbox{myRed}{44.13}\\
Level 1 ($\%$) & 15.59/13.96 & -35.70/-43.58 & 3.80/4.31 & 23.20/26.94 & -153.31/-150.59 & 18.14/17.47 & 9.79/9.48 & 57.94/64.43 & 45.58/48.44\\
Level 2 ($\%$) & 11.97/7.86 & -37.13/-49.84 & 3.40/4.22 & 20.05/26.55 & -163.81/-155.65 & 16.78/16.01 & 7.15/7.15 & 51.63/63.97 & 43.14/47.97\\
Level 3 ($\%$) & 11.34/5.67 & -38.96/-54.28 & 3.13/4.14 & 15.81/26.06 & -176.09/-159.92 & 14.84/14.44 & 5.11/5.11 & 47.55/63.09 & 40.38/47.60\\
Lower Bound ($\%$) & \colorbox{myRed}{23.68} & -43.16 & 0.26 & 1.18 & -330.03 & 0.23 & 0.00 & 35.95 & 10.70\\
\hline
\colorbox{myGreen}{Correlation} & -0.92/-0.96 & -1.00/-1.00 & -0.99/-1.00 & -1.00/-0.99 & -0.99/-0.99  & -0.99/-1.00 & -0.99/-0.99 & -0.99/-0.98 & -1.00/-1.00\\
\colorbox{myGreen}{p-value} & 0.25/0.17 & 0.04*/0.06 & 0.07/0.03* & 0.05*/0.04* & 0.03*/0.03* & 0.06/0.01* & 0.05*/0.02* & 0.07/0.11 & 0.02*/0.04*\\
\hline
\end{tabular}
\end{adjustbox}
\end{center}
\caption{Results of various metrics under the GO FIGURE benchmark on XSum. The symbolic meaning is the same as that in Table \ref{tab:3}.}
\label{tab:4}
\end{table*}

\subsubsection{Interpretability}
We report the interpretability of ClozE by combining the computational procedure of the three metrics in Table \ref{tab:1} and Figure \ref{figure:1}. Obviously, our ClozE framework has the same procedure as QA-based metrics, i.e., extracting factual factors for comparison. This fact indicates that our ClozE has the same interpretability as QA-based metrics, which is stronger than that of NLI-based metrics.

\subsubsection{Convenience}
We believe that the convenience of the metric is as important as the aspects above. To simplify the convenience measurement, we evaluate the inconvenience of the metrics instead of the convenience. We give the scores directly through the number of models involved in the metrics. As shown in Inc. section in Table \ref{tab:2}, most of the QA-metrics have two or more models, while others, including our ClozE, require only one model.

\subsection{Ablation Study}
In Table \ref{tab:2}, we show the main ablation experiment results, and in Table \ref{tab:6} in Appendix B, we present the full ablation experimental results.  
\paragraph{Granularity} In terms of granularity, there is not much difference between summary-level and sentence-level. We believe that the main reason for this is that the candidate summary sentences in the dataset are not very related to each other. Therefore, neither the single-sentence input nor the whole candidate summary input affects the accuracy of the sentence-level cloze.

\paragraph{Factual Factors Extractor} We used two factual factors extractors, \verb|en_core_web_sm| (\verb|SM|) and \verb|en_core_web_trf| (\verb|TRF|) respectively. \verb|SM| has less time spent and, relatively, it extracts a much lower number and accuracy of factual factors than \verb|TRF|. Due to this, we can observe that the time spent on the results with \verb|SM| is much less than that with \verb|TRF|, but the correlation is also basically lower than that with \verb|TRF|.

\paragraph{Confidence} We also conducted experiments on the Confidence proposed in \S \ref{sec:ab}. Because it is represented simply by the generation probabilities of factual factors, it has no impact on the computational speed of ClozE. Despite its simple design, it is also a certain effect enhancement for ClozE. It can help ClozE discard some uncertain cloze results to improve the accuracy of evaluating factual consistency.

\subsection{Case Study}
\label{sec:Case Study}

\begin{table*}[ht]
\footnotesize
\renewcommand\arraystretch{1.2}
\begin{tabularx}{\textwidth}{lX}
\hline
\multicolumn{2}{c}{\bf Example \#1}\\
\hline
\textbf{Document} & […] The patriots beat seattle seahawks 28 - 24 to win the nfl super bowl in glendale, arizona with brady named as mvp for the third time in his fourth super bowl victory. Although brady wasn't able to visit the white house with […] He joins the likes of joe montana and terry bradshaw on four rings, and is now the all-time leader with 12 super bowl touchdown passes and completions on 37. […]\\
\hline
\textbf{Summary} & The new england patriots beat \textcolor{LimeGreen}{seattle seahawks} 28-24 to win super bowl xlix in glendale, \textcolor{LimeGreen}{arizona} with brady named as mvp for the third time in his fourth super bowl victory. [...] \textcolor{red}{The president} was a senator with 12 super bowl touchdown passes and completions on 37.\\
\hline
\textbf{Masked Summary} & The new england patriots beat [MASK] 28-24 to win super bowl xlix in glendale, [MASK] with brady named as mvp for the third time in his fourth super bowl victory. [MASK] was a senator with 12 super bowl touchdown passes and completions on 37.\\
\hline
\textbf{Correct Answers} & \textcolor{LimeGreen}{seattle seahawks}{ }($1.0$),{ }\textcolor{LimeGreen}{arizona}{ }($1.0$),{ }\textcolor{LimeGreen}{brady}{ }($0.0$) \\
\hline
\textbf{Cloze Results} & \textcolor{LimeGreen}{seattle seahawks}{ }($1.0$),{ }\textcolor{LimeGreen}{arizona}{ }($1.0$),{ }\textcolor{LimeGreen}{brady}{ }($0.0$) \\
\hline
\multicolumn{2}{c}{\bf Example \#2} \\
\hline
\textbf{Document} & […] Nominations are open for cnn heroes 2015. A decade - long civil war had just ended in the country, and doyne witnessed its effects firsthand. She met women and children who were suffering, struggling to survive. ` `it changed me,'' said doyne, now 28. […] Today, kopila - - which means `` flower bud'' in nepali - - is home to about 50 children, from infants to teenagers. […]\\
\hline
\textbf{Summary} & Nominations are open for \textcolor{LimeGreen}{cnn} heroes 2015. Doyne, nepal, met women and children in \textcolor{red}{nepal}. \textcolor{red}{The school} is home to about 50 children.\\
\hline
\textbf{Masked Summary} & Nominations are open for [MASK] heroes 2015. Doyne, nepal, met women and children in [MASK]. [MASK] is home to about 50 children.\\
\hline
\textbf{Correct Answers} & \textcolor{LimeGreen}{cnn}{ }($1.0$),{ }\textcolor{LimeGreen}{[None]}{ }($0.0$),{ }\textcolor{LimeGreen}{kopila}{ }($0.0$) \\
\hline
\textbf{Cloze Results} & \textcolor{LimeGreen}{cnn}{ }($1.0$),{ }\textcolor{red}{maggie do}{ }($0.0$),{ }\textcolor{red}{doyne}{ }($0.0$) \\
\hline
\multicolumn{2}{c}{\bf Example \#3} \\
\hline
\textbf{Document} & […] Cameron splayed his legs, he smiled on cue and when he swivelled his huge eyes direct to camera to repeatedly use the phrase `if i am prime minister, it was […] Miliband's coaching team will be patting themselves on the back and declaring a win following the debate. […]\\
\hline
\textbf{Summary} & \textcolor{red}{Nicola sturgeon} is a prime minister and miliband's coaching team. [...] David cameron is a coaching team for \textcolor{red}{cameron}.\\
\hline
\textbf{Masked Summary} & [MASK] is a prime minister and miliband's coaching team. [...] David cameron is a coaching team for [MASK].\\
\hline
\textbf{Correct Answers} & \textcolor{LimeGreen}{cameron}{ }($0.0$),{ }\textcolor{LimeGreen}{miliband}{ }($0.0$) \\
\hline
\textbf{Cloze Results} & \textcolor{red}{m cameronons}{ }($0.0$),{ }\textcolor{LimeGreen}{miliband}{ }($0.0$) \\
\hline
\end{tabularx}
\caption{Examples generated by ClozE on the QAGS-CNN/DM dataset. Where \textcolor{LimeGreen}{green} words indicate correct factual factors, \textcolor{red}{red} ones indicate incorrect factual factors, and \textcolor{LimeGreen}{[None]} indicates that it cannot be answered. The (*) after each factual factor is the similarity score obtained by Eq.2.} 
\label{tab:5}
\end{table*}

In Table \ref{tab:5}, we present three examples, each of which contains document, summary, masked summary, correct answers and cloze results with similarity scores. Obviously, there are only two cases for the cloze results, which are filling in correct factual factors and incorrect ones.

On most of the examples, ClozE fills in the same factual factors when the masked ones are correct, which is a common phenomenon throughout ClozE and one of the reasons for its high performance. Meanwhile, we also note that ClozE replaces some incorrect factual factors with correct ones, such as \verb|"The president"| $\rightarrow$ \verb|"brady"| on the first example and \verb|"cameron"| $\rightarrow$ \verb|"miliband"| on the third example. This is a delightful surprise, indicating that ClozE may be able to correct the wrong factual factors. 

As shown on the second and third examples, ClozE sometimes fills in incorrect answers when the masked factual factors are incorrect either. We believe that the major reason is due to the limitations of autoencoding pre-trained models, which are weak on non-fixed-length generating tasks \shortcite{du2022glm}. 

However, this issue does not lead to poor performance in factual consistency evaluation. As shown in Eq.\ref{equation:2}, the ClozE score is calculated by comparing the factual factors in the summary with those filled in by the cloze model. This means that even if a factual factor generated by the cloze model is wrong, we can still obtain a precise ClozE score, which is verified on the second and third examples.

\section{Discussion}
\label{sec:Discussion}

In this section, we further discuss three important facets of ClozE, including the balance of speed and performance, the optimization of metric and location of factual errors.

\begin{figure}[t]
	\subfloat[]{
	    \includegraphics[width = 0.48\linewidth]{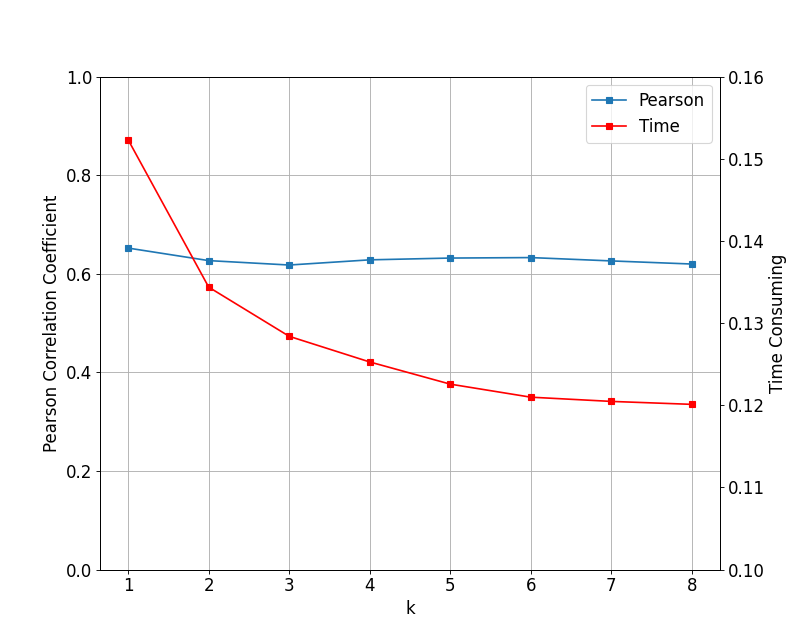}
	    }
	\subfloat[]{
	    \includegraphics[width = 0.52\linewidth]{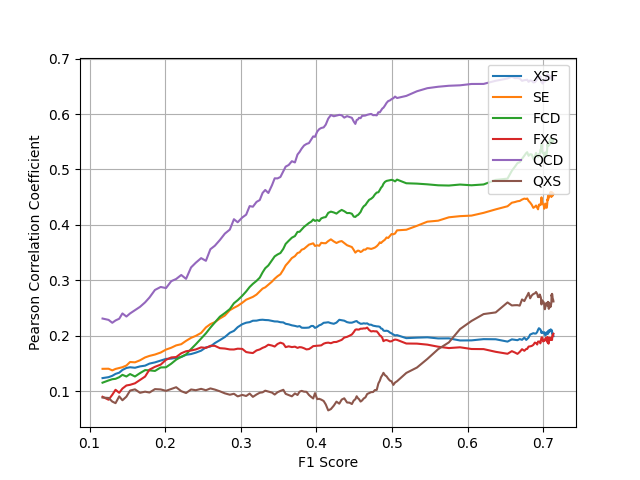}
	    }\\
	\caption{(a){ }The trend of ClozE evaluation speed and correlation with human judgments given different $k$ settings on QAGS-CNN/DM.{ }(b){ }The trend of cloze accuracy (F1 Score) and correlation with human judgments on different datasets.}
	\label{figure:3}
\end{figure}

\paragraph{Balance of Speed and Performance}
The number of questions that can be answered at once is critical to the speed of evaluation. The low speed of QA-based metrics is due to the limitation of answering one question at once. And for NLI-based metrics, it can be considered as answering infinite questions at once, which improves the speed of evaluation greatly. Following this characteristic of NLI-based metrics, our ClozE can mask $k$ factual factors at once (\S \ref{sec:k}), which is equivalent to simultaneously generate and answer $k$ questions. We present the performance and speed of ClozE with different $k$ in Figure \ref{figure:3}(a) and more results on different datasets in Figure \ref{figure:4} in Appendix B. We use summary-level granularity for a larger range of $k$ and greedily mask the $k$ consecutive factual factors together one by one. We observe that the time (in seconds) taken by ClozE decreases linearly as the number $k$ increases. The performance of the model is the highest when $k=1$ and it decreases slowly when $k>1$, reflected by the Pearson correlation coefficient. The reason is that there may be causal relationships between factual factors, and when we replace them with \verb|[MASK]| at the same time, the cloze model will have a degradation of performance due to the lack of information. In some cases, we can sacrifice a little performance in this way in exchange for faster evaluation.

\paragraph{Optimization of Metric}

One direction of optimization for model-based metrics is to reduce the cumulative errors between modules. The QG and QA modules of QA-based metrics lead to errors accumulating between modules and complicates the study of how to improve the QA-based metrics. In contrast, NLI-metrics alleviate this issue due to the use of only one model. Following this insight, we perform further experiments and analyze on ClozE to explore the correlation between cloze performance and metric performance. Because the performance of cloze depends on the structure of the model as well as the degree of training, we test both Roberta-base and Bart-base model and their checkpoints at different training steps (Figure \ref{figure:3}(b)) and observe a clear upward trend on most of the datasets. Although there is a slight decrease on XSF and FXS, the decrease is not significant and acceptable. Meanwhile, \S 4.6 shows that ClozE does only care whether the generated answers are correct when the factual factors in the summary are correct, which means our method is equivalent to the autoencoding pre-trained models with MLM. Overall, we can assume that the performance of our ClozE can be improved when we have a better autoencoding pre-trained model as the cloze model, which brings great convenience in further improving factual consistency metrics.

\paragraph{Location of Factual Errors} In practice, we usually not only need the metric to give us a score, but also want it to help us analyze where the factual inconsistency errors occur, so as to help us improve our work. As mentioned above, NLI-metrics are lack of interpretability, so it cannot locate where the errors occur. In contrast, our ClozE can naturally locate the incorrect factual factors when it computes scores, which is similar to QA-based metrics. As shown in Table \ref{tab:5}, the factual factors in the summary that differ from the answers generated by the cloze model will be considered as the factual errors.

\section{Conclusion}
In this paper, we propose ClozE -- a cloze-based factual consistency evaluation framework, which leverages the best of both NLI-based and QA-based metrics while avoiding their limitations. It carries the strong interpretability and fast evaluation speed while develops more advantages and the comparisons at all aspects in \S \ref{sec:results} and \S \ref{sec:Discussion} also show its great potential. 
We believe that a faster and effective factual consistency metric can greatly promote the study of factual consisitency in abstractive summarization and hope our findings in this paper will provide insights to future work in all the factual consistency related research. Moreover, we will continue to further research for evaluating more abstractive summarization, higher overall performance and faster speed. 

\acks{This work was supported by the National Natural Science Foundation of China (Grant No. 62176024); the National Key R\&D Program of China (2022ZD01161); Beijing Municipal Science \& Technology Commission [Grant No. Z181100001018035]; Engineering Research Center of Information Networks, Ministry of Education; the Fundamental Research Funds for the Central Universities (2021XD-A01-1).}

\newpage
\appendix

\section*{Appendix A. Details of Benchmarks}
\label{appendix:A}
We will show the details of the benchmarks we use in this section.
\paragraph{XSumFaith} \shortciteA{maynez2020faithfulness} focused on XSum because the golden summaries of this dataset are more abstract and can bring out more of the factual consistency issues of the system summaries. They first asked the annotators to label various types of hallucinations in the system summary and then calculated the factual consistency score of the system summary based on the number of these hallucinations as a reference.
\paragraph{SummEval} \shortciteA{fabbri2021summeval} randomly selected 100 articles from the CNN/DailyMail test set and generated numerous system summaries for crowd-sourced and expert annotation by means of an existing neural summary model. Also, the annotation was done in the same way as \shortciteA{kryscinski2019neural}, and the model output was evaluated according to the four aspects of consistency, coherence, fluency, and relevance. We used the score of consistency as the result of the factual consistency assessment of the systematic summary of the document.
\paragraph{FRANK} \shortciteA{pagnoni2021understanding} included model summaries from CNN/DM and XSum datasets. They annotated each sentence of a summary with numerous error types based on their proposed criteria. In order to obtain the human evaluation score, we consider all error types uniformly and compute the score based on the percentage of errors.
\paragraph{QAGS} \shortciteA{wang2020asking} also took into account the system summaries generated on both the CNNDM and XSum datasets. They collected 3 annotations for each summary and gave a score for a single sentence by voting. Then all sentences were averaged to get the final human evaluation score.
\paragraph{GO FIGURE} \shortciteA{gabriel2021go} differed from previous work on evaluating the factuality of summaries in that they tried to give a criterion to rate the performance of these aforementioned metrics for evaluating factual consistency. They gave five objective conditions: (1) the evaluation score in the generated summary should be greater than that in the completely irrelevant document and less than or equal to that in the golden summary; (2) the evaluation score is negatively correlated with the number of factual errors; (3) the performance is not affected by different factual error types; (4) the performance is not affected by different domain data; (5) the evaluation score is highly correlated with the human judgments.

\newpage
\section*{Appendix B. More Experiments}
\label{appendix:B}
\begin{figure*}
\includegraphics[width=\textwidth]{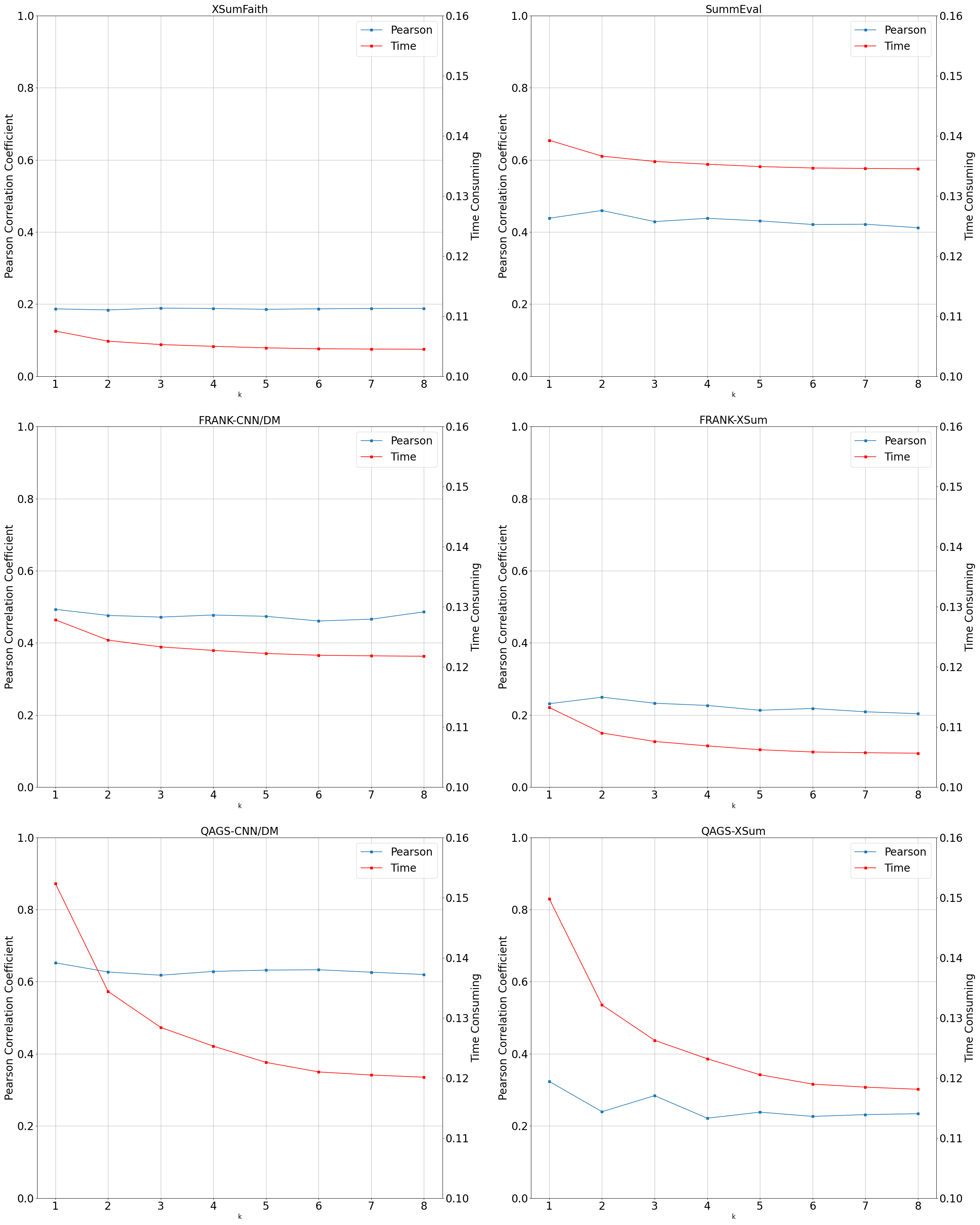}
\caption{The effect of different $k$ settings on ClozE performance on six datasets.}
\label{figure:4}
\end{figure*}

\begin{table*}[ht]
    \begin{center}
        \begin{adjustbox}{max width=1\linewidth}
\begin{tabular}{clccccccc}
\hline
\textbf{Type} & \textbf{Metric} &\textbf{XSF} &\textbf{SE} &\textbf{FCD} &\textbf{FXS} &\textbf{QCD} &\textbf{QXS} &\textbf{Sec./Sum}\\
\hline
\multirow{10}{*}{ClozE-R} & \textbf{summary-level} \\
                         & \verb| +en_core_web_sm| & 16.45 & 44.29 & 49.70 & 21.33 & 65.76 & 27.79 & 0.13 \\
                        & \verb| +confidence| & 17.19 & 45.60 & 49.39 & 21.93 & \textbf{66.30} & 24.97 & 0.13 \\
                        & \textbf{summary-level}\\
                        &\verb| +en_core_web_trf| & 18.67 & 43.81 & 49.31 & 23.15 & 65.24 & \textbf{32.36} & 0.23 \\
                        & \verb| +confidence| & 19.20 & 45.05 & 48.71 & \textbf{23.63} & 65.90 & 29.37 & 0.23 \\
                    & \textbf{sentence-level}\\
                    & \verb| +en_core_web_sm| & 16.45 & 47.17 & \textbf{57.27} & 21.27 & 65.20 & 27.79 & 0.11 \\
                    & \verb| +confidence| & 17.09 & 47.97 & 56.82 & 21.89 & 64.93 & 24.97 & 0.11 \\
                    & \textbf{sentence-level}\\
                    & \verb| +en_core_web_trf| & 18.66 & 47.09 & 55.38 & 23.10 & 66.12 & \textbf{32.36} & 0.21 \\
                    & \verb| +confidence| & 19.22 & \textbf{48.01} & 55.12 & 23.54 & 65.59 & 29.37 & 0.21 \\\hline
                    
\multirow{10}{*}{ClozE-B} & \textbf{summary-level} \\
                        & \verb| +en_core_web_sm| & 20.05 & 38.17 & 45.21 & 19.31 & 63.39 & 20.56 & 0.09 \\
                        & \verb| +confidence| & 19.63 & 37.52 & 43.84 & 20.12 & 64.39 & 21.62 & 0.09 \\
                        & \textbf{summary-level}\\
                        & \verb| +en_core_web_trf| & \textbf{20.47} & 40.08 & 46.74 & 20.38 & 63.72 & 19.60 & 0.19 \\
                        & \verb| +confidence| & 20.46 & 39.52 & 45.47 & 20.86 & 65.13 & 22.16 & 0.19 \\
                        & \textbf{sentence-level}\\
                        & \verb| +en_core_web_sm| & 20.08 & 35.33 & 42.12 & 19.27 & 61.07 & 20.56 & \textbf{0.08} \\
                        & \verb| +confidence| & 19.64 & 34.77 & 40.48 & 20.09 & 60.69 & 21.62 & \textbf{0.08} \\
                        & \textbf{sentence-level} \\
                        & \verb| +en_core_web_trf| & 20.44 & 36.82 & 43.26 & 20.40 & 61.68 & 19.42 & 0.18 \\
                        & \verb| +confidence| & 20.44 & 36.47 & 42.12 & 20.89 & 61.01 & 21.99 & 0.18 \\\hline
\end{tabular}
\end{adjustbox}
\end{center}
\caption{Full ablation experiments on six datasets. We also set the parameters $k = 1$ and $\alpha=\beta=0.5$.}
\label{tab:6}
\end{table*}

\newpage

\begin{thebibliography}{}

\bibitem[\protect\BCAY{Banerjee\ \BBA\ Lavie}{Banerjee\ \BBA\
  Lavie}{2005}]{banerjee2005meteor}
Banerjee, S.\BBACOMMA\  \BBA\ Lavie, A. \BBOP2005\BBCP.
\newblock \BBOQ Meteor: An automatic metric for mt evaluation with improved
  correlation with human judgments\BBCQ\
\newblock In {\Bem Proceedings of the acl workshop on intrinsic and extrinsic
  evaluation measures for machine translation and/or summarization}, \BPGS\
  65--72.

\bibitem[\protect\BCAY{Barrantes, Herudek,\ \BBA\ Wang}{Barrantes
  et~al.}{2020}]{barrantes2020adversarial}
Barrantes, M., Herudek, B., \BBA\ Wang, R. \BBOP2020\BBCP.
\newblock \BBOQ Adversarial nli for factual correctness in text summarisation
  models\BBCQ\
\newblock {\Bem arXiv preprint arXiv:2005.11739}.

\bibitem[\protect\BCAY{Cao, Dong, Wu,\ \BBA\ Cheung}{Cao
  et~al.}{2020}]{cao2020factual}
Cao, M., Dong, Y., Wu, J., \BBA\ Cheung, J. C.~K. \BBOP2020\BBCP.
\newblock \BBOQ Factual error correction for abstractive summarization
  models\BBCQ\
\newblock In {\Bem Proceedings of the 2020 Conference on Empirical Methods in
  Natural Language Processing (EMNLP)}, \BPGS\ 6251--6258.

\bibitem[\protect\BCAY{Cao, Wei, Li,\ \BBA\ Li}{Cao
  et~al.}{2018}]{cao2018faithful}
Cao, Z., Wei, F., Li, W., \BBA\ Li, S. \BBOP2018\BBCP.
\newblock \BBOQ Faithful to the original: Fact aware neural abstractive
  summarization\BBCQ\
\newblock In {\Bem thirty-second AAAI conference on artificial intelligence}.

\bibitem[\protect\BCAY{Deutsch\ \BBA\ Roth}{Deutsch\ \BBA\
  Roth}{2021}]{deutsch2021understanding}
Deutsch, D.\BBACOMMA\  \BBA\ Roth, D. \BBOP2021\BBCP.
\newblock \BBOQ Understanding the extent to which content quality metrics
  measure the information quality of summaries\BBCQ\
\newblock In {\Bem Proceedings of the 25th Conference on Computational Natural
  Language Learning}, \BPGS\ 300--309.

\bibitem[\protect\BCAY{Du, Qian, Liu, Ding, Qiu, Yang,\ \BBA\ Tang}{Du
  et~al.}{2022}]{du2022glm}
Du, Z., Qian, Y., Liu, X., Ding, M., Qiu, J., Yang, Z., \BBA\ Tang, J.
  \BBOP2022\BBCP.
\newblock \BBOQ Glm: General language model pretraining with autoregressive
  blank infilling\BBCQ\
\newblock In {\Bem Proceedings of the 60th Annual Meeting of the Association
  for Computational Linguistics (Volume 1: Long Papers)}, \BPGS\ 320--335.

\bibitem[\protect\BCAY{Durmus, He,\ \BBA\ Diab}{Durmus
  et~al.}{2020}]{durmus2020feqa}
Durmus, E., He, H., \BBA\ Diab, M. \BBOP2020\BBCP.
\newblock \BBOQ Feqa: A question answering evaluation framework for
  faithfulness assessment in abstractive summarization\BBCQ\
\newblock In {\Bem Proceedings of the 58th Annual Meeting of the Association
  for Computational Linguistics}, \BPGS\ 5055--5070.

\bibitem[\protect\BCAY{Fabbri, Kry{\'s}ci{\'n}ski, McCann, Xiong, Socher,\
  \BBA\ Radev}{Fabbri et~al.}{2021a}]{fabbri2021summeval}
Fabbri, A.~R., Kry{\'s}ci{\'n}ski, W., McCann, B., Xiong, C., Socher, R., \BBA\
  Radev, D. \BBOP2021a\BBCP.
\newblock \BBOQ Summeval: Re-evaluating summarization evaluation\BBCQ\
\newblock {\Bem Transactions of the Association for Computational Linguistics},
  {\Bem 9}, 391--409.

\bibitem[\protect\BCAY{Fabbri, Wu, Liu,\ \BBA\ Xiong}{Fabbri
  et~al.}{2021b}]{fabbri2021qafacteval}
Fabbri, A.~R., Wu, C.-S., Liu, W., \BBA\ Xiong, C. \BBOP2021b\BBCP.
\newblock \BBOQ Qafacteval: Improved qa-based factual consistency evaluation
  for summarization\BBCQ\
\newblock {\Bem arXiv preprint arXiv:2112.08542}.

\bibitem[\protect\BCAY{Falke, Ribeiro, Utama, Dagan,\ \BBA\ Gurevych}{Falke
  et~al.}{2019}]{falke2019ranking}
Falke, T., Ribeiro, L.~F., Utama, P.~A., Dagan, I., \BBA\ Gurevych, I.
  \BBOP2019\BBCP.
\newblock \BBOQ Ranking generated summaries by correctness: An interesting but
  challenging application for natural language inference\BBCQ\
\newblock In {\Bem Proceedings of the 57th Annual Meeting of the Association
  for Computational Linguistics}, \BPGS\ 2214--2220.

\bibitem[\protect\BCAY{Gabriel, Celikyilmaz, Jha, Choi,\ \BBA\ Gao}{Gabriel
  et~al.}{2021}]{gabriel2021go}
Gabriel, S., Celikyilmaz, A., Jha, R., Choi, Y., \BBA\ Gao, J. \BBOP2021\BBCP.
\newblock \BBOQ Go figure: A meta evaluation of factuality in
  summarization\BBCQ\
\newblock In {\Bem Findings of the Association for Computational Linguistics:
  ACL-IJCNLP 2021}, \BPGS\ 478--487.

\bibitem[\protect\BCAY{Goyal\ \BBA\ Durrett}{Goyal\ \BBA\
  Durrett}{2021}]{goyal2021annotating}
Goyal, T.\BBACOMMA\  \BBA\ Durrett, G. \BBOP2021\BBCP.
\newblock \BBOQ Annotating and modeling fine-grained factuality in
  summarization\BBCQ\
\newblock In {\Bem Proceedings of the 2021 Conference of the North American
  Chapter of the Association for Computational Linguistics: Human Language
  Technologies}, \BPGS\ 1449--1462.

\bibitem[\protect\BCAY{Hermann, Kocisky, Grefenstette, Espeholt, Kay,
  Suleyman,\ \BBA\ Blunsom}{Hermann et~al.}{2015}]{hermann2015teaching}
Hermann, K.~M., Kocisky, T., Grefenstette, E., Espeholt, L., Kay, W., Suleyman,
  M., \BBA\ Blunsom, P. \BBOP2015\BBCP.
\newblock \BBOQ Teaching machines to read and comprehend\BBCQ\
\newblock {\Bem Advances in neural information processing systems}, {\Bem 28}.

\bibitem[\protect\BCAY{Kenton\ \BBA\ Toutanova}{Kenton\ \BBA\
  Toutanova}{2019}]{kenton2019bert}
Kenton, J. D. M.-W.~C.\BBACOMMA\  \BBA\ Toutanova, L.~K. \BBOP2019\BBCP.
\newblock \BBOQ Bert: Pre-training of deep bidirectional transformers for
  language understanding\BBCQ\
\newblock In {\Bem Proceedings of NAACL-HLT}, \BPGS\ 4171--4186.

\bibitem[\protect\BCAY{Kingma\ \BBA\ Ba}{Kingma\ \BBA\
  Ba}{2014}]{kingma2014adam}
Kingma, D.~P.\BBACOMMA\  \BBA\ Ba, J. \BBOP2014\BBCP.
\newblock \BBOQ Adam: A method for stochastic optimization\BBCQ\
\newblock {\Bem arXiv preprint arXiv:1412.6980}.

\bibitem[\protect\BCAY{Koto, Baldwin,\ \BBA\ Lau}{Koto
  et~al.}{2022}]{koto2022ffci}
Koto, F., Baldwin, T., \BBA\ Lau, J.~H. \BBOP2022\BBCP.
\newblock \BBOQ Ffci: A framework for interpretable automatic evaluation of
  summarization\BBCQ\
\newblock {\Bem Journal of Artificial Intelligence Research}, {\Bem 73},
  1553--1607.

\bibitem[\protect\BCAY{Kry{\'s}ci{\'n}ski, Keskar, McCann, Xiong,\ \BBA\
  Socher}{Kry{\'s}ci{\'n}ski et~al.}{2019}]{kryscinski2019neural}
Kry{\'s}ci{\'n}ski, W., Keskar, N.~S., McCann, B., Xiong, C., \BBA\ Socher, R.
  \BBOP2019\BBCP.
\newblock \BBOQ Neural text summarization: A critical evaluation\BBCQ\
\newblock In {\Bem Proceedings of the 2019 Conference on Empirical Methods in
  Natural Language Processing and the 9th International Joint Conference on
  Natural Language Processing (EMNLP-IJCNLP)}, \BPGS\ 540--551.

\bibitem[\protect\BCAY{Kry{\'s}ci{\'n}ski, McCann, Xiong,\ \BBA\
  Socher}{Kry{\'s}ci{\'n}ski et~al.}{2020}]{kryscinski2020evaluating}
Kry{\'s}ci{\'n}ski, W., McCann, B., Xiong, C., \BBA\ Socher, R. \BBOP2020\BBCP.
\newblock \BBOQ Evaluating the factual consistency of abstractive text
  summarization\BBCQ\
\newblock In {\Bem Proceedings of the 2020 Conference on Empirical Methods in
  Natural Language Processing (EMNLP)}, \BPGS\ 9332--9346.

\bibitem[\protect\BCAY{Laban, Schnabel, Bennett,\ \BBA\ Hearst}{Laban
  et~al.}{2022}]{laban2022summac}
Laban, P., Schnabel, T., Bennett, P.~N., \BBA\ Hearst, M.~A. \BBOP2022\BBCP.
\newblock \BBOQ Summac: Re-visiting nli-based models for inconsistency
  detection in summarization\BBCQ\
\newblock {\Bem Transactions of the Association for Computational Linguistics},
  {\Bem 10}, 163--177.

\bibitem[\protect\BCAY{Lewis, Liu, Goyal, Ghazvininejad, Mohamed, Levy,
  Stoyanov,\ \BBA\ Zettlemoyer}{Lewis et~al.}{2020}]{lewis2020bart}
Lewis, M., Liu, Y., Goyal, N., Ghazvininejad, M., Mohamed, A., Levy, O.,
  Stoyanov, V., \BBA\ Zettlemoyer, L. \BBOP2020\BBCP.
\newblock \BBOQ Bart: Denoising sequence-to-sequence pre-training for natural
  language generation, translation, and comprehension\BBCQ\
\newblock In {\Bem Proceedings of the 58th Annual Meeting of the Association
  for Computational Linguistics}, \BPGS\ 7871--7880.

\bibitem[\protect\BCAY{Lin}{Lin}{2004}]{lin2004rouge}
Lin, C.-Y. \BBOP2004\BBCP.
\newblock \BBOQ Rouge: A package for automatic evaluation of summaries\BBCQ\
\newblock In {\Bem Text summarization branches out}, \BPGS\ 74--81.

\bibitem[\protect\BCAY{Liu, Yuan, Fu, Jiang, Hayashi,\ \BBA\ Neubig}{Liu
  et~al.}{2021}]{liu2021pre}
Liu, P., Yuan, W., Fu, J., Jiang, Z., Hayashi, H., \BBA\ Neubig, G.
  \BBOP2021\BBCP.
\newblock \BBOQ Pre-train, prompt, and predict: A systematic survey of
  prompting methods in natural language processing\BBCQ\
\newblock {\Bem arXiv preprint arXiv:2107.13586}.

\bibitem[\protect\BCAY{Liu, Ott, Goyal, Du, Joshi, Chen, Levy, Lewis,
  Zettlemoyer,\ \BBA\ Stoyanov}{Liu et~al.}{2019}]{liu2019roberta}
Liu, Y., Ott, M., Goyal, N., Du, J., Joshi, M., Chen, D., Levy, O., Lewis, M.,
  Zettlemoyer, L., \BBA\ Stoyanov, V. \BBOP2019\BBCP.
\newblock \BBOQ Roberta: A robustly optimized bert pretraining approach\BBCQ\
\newblock {\Bem arXiv preprint arXiv:1907.11692}.

\bibitem[\protect\BCAY{Maynez, Narayan, Bohnet,\ \BBA\ McDonald}{Maynez
  et~al.}{2020}]{maynez2020faithfulness}
Maynez, J., Narayan, S., Bohnet, B., \BBA\ McDonald, R. \BBOP2020\BBCP.
\newblock \BBOQ On faithfulness and factuality in abstractive
  summarization\BBCQ\
\newblock In {\Bem Proceedings of the 58th Annual Meeting of the Association
  for Computational Linguistics}, \BPGS\ 1906--1919.

\bibitem[\protect\BCAY{Nan, Santos, Zhu, Ng, McKeown, Nallapati, Zhang, Wang,
  Arnold,\ \BBA\ Xiang}{Nan et~al.}{2021}]{nan2021improving}
Nan, F., Santos, C. N.~d., Zhu, H., Ng, P., McKeown, K., Nallapati, R., Zhang,
  D., Wang, Z., Arnold, A.~O., \BBA\ Xiang, B. \BBOP2021\BBCP.
\newblock \BBOQ Improving factual consistency of abstractive summarization via
  question answering\BBCQ\
\newblock {\Bem arXiv preprint arXiv:2105.04623}.

\bibitem[\protect\BCAY{Narayan, Cohen,\ \BBA\ Lapata}{Narayan
  et~al.}{2018}]{narayan2018don}
Narayan, S., Cohen, S., \BBA\ Lapata, M. \BBOP2018\BBCP.
\newblock \BBOQ Don't give me the details, just the summary! topic-aware
  convolutional neural networks for extreme summarization\BBCQ\
\newblock In {\Bem 2018 Conference on Empirical Methods in Natural Language
  Processing}, \BPGS\ 1797--1807. Association for Computational Linguistics.

\bibitem[\protect\BCAY{Pagnoni, Balachandran,\ \BBA\ Tsvetkov}{Pagnoni
  et~al.}{2021}]{pagnoni2021understanding}
Pagnoni, A., Balachandran, V., \BBA\ Tsvetkov, Y. \BBOP2021\BBCP.
\newblock \BBOQ Understanding factuality in abstractive summarization with
  frank: A benchmark for factuality metrics\BBCQ\
\newblock In {\Bem Proceedings of the 2021 Conference of the North American
  Chapter of the Association for Computational Linguistics: Human Language
  Technologies}, \BPGS\ 4812--4829.

\bibitem[\protect\BCAY{Papineni, Roukos, Ward,\ \BBA\ Zhu}{Papineni
  et~al.}{2002}]{papineni2002bleu}
Papineni, K., Roukos, S., Ward, T., \BBA\ Zhu, W.-J. \BBOP2002\BBCP.
\newblock \BBOQ Bleu: a method for automatic evaluation of machine
  translation\BBCQ\
\newblock In {\Bem Proceedings of the 40th annual meeting of the Association
  for Computational Linguistics}, \BPGS\ 311--318.

\bibitem[\protect\BCAY{Qasemi, Kezar, Pujara,\ \BBA\ Szekely}{Qasemi
  et~al.}{2022}]{qasemi2022evaluating}
Qasemi, E., Kezar, L., Pujara, J., \BBA\ Szekely, P. \BBOP2022\BBCP.
\newblock \BBOQ Evaluating machine common sense via cloze testing\BBCQ\
\newblock {\Bem arXiv preprint arXiv:2201.07902}.

\bibitem[\protect\BCAY{Rei, Stewart, Farinha,\ \BBA\ Lavie}{Rei
  et~al.}{2020}]{rei2020comet}
Rei, R., Stewart, C., Farinha, A.~C., \BBA\ Lavie, A. \BBOP2020\BBCP.
\newblock \BBOQ Comet: A neural framework for mt evaluation\BBCQ\
\newblock In {\Bem Proceedings of the 2020 Conference on Empirical Methods in
  Natural Language Processing (EMNLP)}, \BPGS\ 2685--2702.

\bibitem[\protect\BCAY{Scialom, Dray, Gallinari, Lamprier, Piwowarski,
  Staiano,\ \BBA\ Wang}{Scialom et~al.}{2021}]{scialom2021questeval}
Scialom, T., Dray, P.-A., Gallinari, P., Lamprier, S., Piwowarski, B., Staiano,
  J., \BBA\ Wang, A. \BBOP2021\BBCP.
\newblock \BBOQ Questeval: Summarization asks for fact-based evaluation\BBCQ\
\newblock In {\Bem Proceedings of the 2021 Conference on Empirical Methods in
  Natural Language Processing}, \BPGS\ 6594--6604. Association for
  Computational Linguistics.

\bibitem[\protect\BCAY{Scialom, Lamprier, Piwowarski,\ \BBA\ Staiano}{Scialom
  et~al.}{2019}]{scialom2019answers}
Scialom, T., Lamprier, S., Piwowarski, B., \BBA\ Staiano, J. \BBOP2019\BBCP.
\newblock \BBOQ Answers unite! unsupervised metrics for reinforced
  summarization models\BBCQ\
\newblock In {\Bem 2019 Conference on Empirical Methods in Natural Language
  Processing and the 9th International Joint Conference on Natural Language
  Processing (EMNLP-IJCNLP)}, \BPGS\ 3237--3247. Association for Computational
  Linguistics.

\bibitem[\protect\BCAY{Sellam, Das,\ \BBA\ Parikh}{Sellam
  et~al.}{2020}]{sellam2020bleurt}
Sellam, T., Das, D., \BBA\ Parikh, A. \BBOP2020\BBCP.
\newblock \BBOQ Bleurt: Learning robust metrics for text generation\BBCQ\
\newblock In {\Bem Proceedings of the 58th Annual Meeting of the Association
  for Computational Linguistics}, \BPGS\ 7881--7892.

\bibitem[\protect\BCAY{Taylor}{Taylor}{1953}]{taylor1953cloze}
Taylor, W.~L. \BBOP1953\BBCP.
\newblock \BBOQ “cloze procedure”: A new tool for measuring
  readability\BBCQ\
\newblock {\Bem Journalism quarterly}, {\Bem 30\/}(4), 415--433.

\bibitem[\protect\BCAY{Vasilyev, Dharnidharka,\ \BBA\ Bohannon}{Vasilyev
  et~al.}{2020}]{vasilyev2020fill}
Vasilyev, O., Dharnidharka, V., \BBA\ Bohannon, J. \BBOP2020\BBCP.
\newblock \BBOQ Fill in the blanc: Human-free quality estimation of document
  summaries\BBCQ\
\newblock In {\Bem Proceedings of the First Workshop on Evaluation and
  Comparison of NLP Systems}, \BPGS\ 11--20.

\bibitem[\protect\BCAY{Vaswani, Shazeer, Parmar, Uszkoreit, Jones, Gomez,
  Kaiser,\ \BBA\ Polosukhin}{Vaswani et~al.}{2017}]{vaswani2017attention}
Vaswani, A., Shazeer, N., Parmar, N., Uszkoreit, J., Jones, L., Gomez, A.~N.,
  Kaiser, {\L}., \BBA\ Polosukhin, I. \BBOP2017\BBCP.
\newblock \BBOQ Attention is all you need\BBCQ\
\newblock {\Bem Advances in neural information processing systems}, {\Bem 30}.

\bibitem[\protect\BCAY{Wang, Cho,\ \BBA\ Lewis}{Wang
  et~al.}{2020}]{wang2020asking}
Wang, A., Cho, K., \BBA\ Lewis, M. \BBOP2020\BBCP.
\newblock \BBOQ Asking and answering questions to evaluate the factual
  consistency of summaries\BBCQ\
\newblock In {\Bem Proceedings of the 58th Annual Meeting of the Association
  for Computational Linguistics}, \BPGS\ 5008--5020.

\bibitem[\protect\BCAY{Wolf, Debut, Sanh, Chaumond, Delangue, Moi, Cistac,
  Rault, Louf, Funtowicz, et~al.}{Wolf et~al.}{2019}]{wolf2019huggingface}
Wolf, T., Debut, L., Sanh, V., Chaumond, J., Delangue, C., Moi, A., Cistac, P.,
  Rault, T., Louf, R., Funtowicz, M., et~al. \BBOP2019\BBCP.
\newblock \BBOQ Huggingface's transformers: State-of-the-art natural language
  processing\BBCQ\
\newblock {\Bem arXiv preprint arXiv:1910.03771}.

\bibitem[\protect\BCAY{Zhang, Kishore, Wu, Weinberger,\ \BBA\ Artzi}{Zhang
  et~al.}{2019}]{zhang2019bertscore}
Zhang, T., Kishore, V., Wu, F., Weinberger, K.~Q., \BBA\ Artzi, Y.
  \BBOP2019\BBCP.
\newblock \BBOQ Bertscore: Evaluating text generation with bert\BBCQ\
\newblock In {\Bem International Conference on Learning Representations}.

\end{thebibliography}

\end{document}